\documentclass[letterpaper, 10 pt, conference]{ieeeconf}  % Comment this line out if you need a4paper

\IEEEoverridecommandlockouts                              % This command is only needed if 
\usepackage{graphicx}
\usepackage{hyperref}
\usepackage{amsmath}
\usepackage{cuted}
\usepackage{capt-of}
\usepackage{booktabs}
\usepackage{makecell}
\usepackage{multirow}
\usepackage{tabularx}
\usepackage{balance}
\usepackage{amssymb}

\title{\LARGE \bf
Bridging the Morphology Gap: Adapting VLA Models to 
Dexterous Manipulation via Intent-Conditioned Fine-Tuning 
}

\author{Chuanke Pang$^{1}$, Junyi Huang$^{1}$, Jianye Fang$^{2}$, Zhijun Zhao$^{3}$, Yaobing Wang$^{3}$, Kun Xu$^{1*}$ and Xilun Ding$^{1}$% <-this % stops a space
\thanks{$^{*}$Corresponding author: Kun Xu (e-mail: \texttt{xk007@buaa.edu.cn}).}%
\thanks{$^{1}$Robotics Institute, Beihang University, Beijing, China}%
\thanks{$^{2}$University of Jinan, Shandong, China}%
\thanks{$^{3}$China Academy of Space Technology, Beijing, China}%
}

\begin{document}

\maketitle % 正常生成标题，千万不要删！
\thispagestyle{empty}
\pagestyle{empty}

% \begin{strip}
%     \begin{center}
%     \vspace{-47pt}
%     \includegraphics[width=0.98\linewidth]{image/lift_pick_place_intent_preview.pdf}
%     \captionof{figure}{\textbf{Intent-conditioned rollout visualization.}
%     Lift and Pick \& Place task-space rollouts are shown on the left and
%     right, respectively; blue, orange, and green denote approach,
%     grasp/release, and lift/transfer. Translucent paths visualize 50
%     conditional action-chunk samples per stage. The center
%     panels show the aperture-derived intent interface, one dexterous
%     realization, and phase-aligned intent traces.}
%     \label{fig:banner_results}
%     \vspace{0pt}
%   \end{center}
% \end{strip}

%%%%%%%%%%%%%%%%%%%%%%%%%%%%%%%%%%%%%%%%%%%%%%%%%%%%%%%%%%%%%%%%%%%%%%%%%%%%%%%%
\begin{abstract}Pretrained vision-language-action (VLA) models offer semantic and spatial priors for robot manipulation, but their parallel-gripper interfaces leave multi-finger contact coordination implicit. Adapting these models requires both timely grasp transitions and embodiment-specific hand trajectories. We introduce \textbf{InDex}, an intent-conditioned framework that connects these two requirements. A normalized aperture signal, derived from retargeted demonstrations, supervises a scalar grasp intent while preserving full hand-action supervision. Stage 1 learns synchronized end-effector--intent chunks; Stage 2 conditions a diffusion hand decoder on these predictions, visual-language context, and proprioception. Sequential optimization establishes the upstream interface before learning dexterous control with low-rate co-adaptation. Across four simulated tasks, InDex with $\pi_{0.5}$ achieves 85.8\% mean task success, improving the native policy by 35.5 percentage points, and reaches 69.0\% on held-out objects with similar geometry without fine-tuning or recalibration. Improvements also hold across three VLA backbones and separately trained real-world policies, where mean success increases from 71.0\% to 83.0\%. Component ablations and timing interventions support the value of explicit grasp-intent conditioning and expressive hand generation. These results identify arm--hand coordination as an actionable bottleneck in adapting parallel-gripper priors to dexterous manipulation.\end{abstract}

%%%%%%%%%%%%%%%%%%%%%%%%%%%%%%%%%%%%%%%%%%%%%%%%%%%%%%%%%%%%%%%%%%%%%%%%%%%%%%%%
\section{Introduction}Vision-language-action (VLA) models learn semantic and spatial priors from large robot datasets~\cite{ref_pi0}, many collected with parallel grippers~\cite{ref_droid,ref_openx}. Extending these priors to dexterous hands could reduce the burden of learning manipulation from limited embodiment-specific demonstrations. However, reaching an object is only the first step: successful manipulation also requires finger pre-shaping, timely contact, grasp stabilization, and coordinated transport or insertion.\par The central challenge is the interface between task-space motion and multi-finger control. Expanding the action vector to include hand joints permits dexterous commands but leaves the timing of contact implicit. Moreover, multiple hand trajectories can realize the same approach and grasp, making deterministic regression vulnerable to averaging incompatible configurations. The resulting gap is therefore both temporal and geometric: a policy must coordinate \emph{when to establish contact} with \emph{how the hand realizes it}. An explicit transition signal must preserve, rather than replace, the hand's configuration-dependent dexterity.\begin{figure}[t]
\centering
\includegraphics[width=0.96\columnwidth]{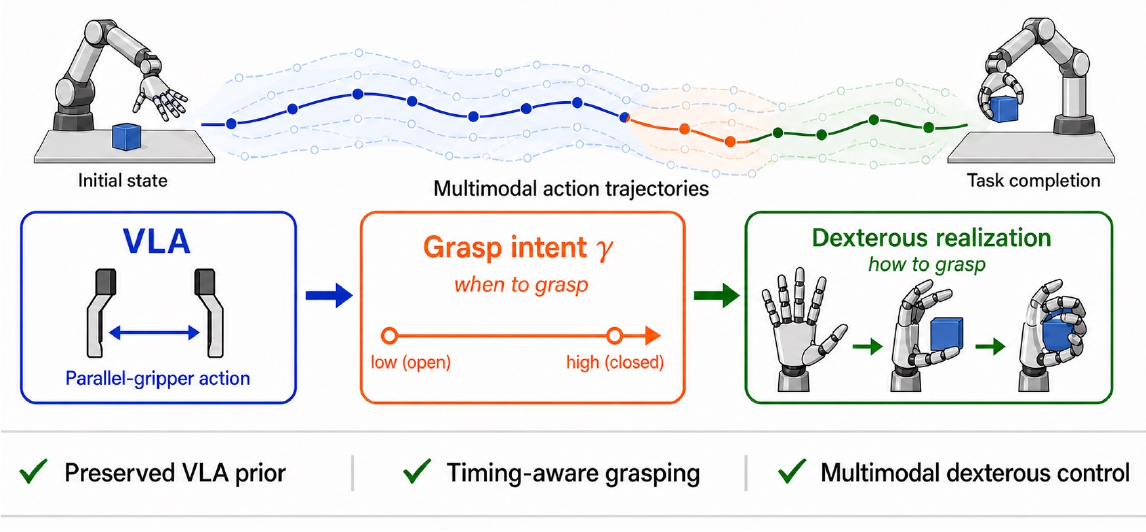}
    \caption{\textbf{Overview of InDex.} Grasp intent links pretrained parallel-gripper VLA priors to diffusion-based dexterous control for coordinated arm--hand manipulation.}
	\label{fig:concept_banner}
\end{figure}

InDex (\textbf{I}ntent-conditioned \textbf{n}ested \textbf{Dex}terous manipulation; Fig.~\ref{fig:concept_banner}) addresses this gap with a normalized grasp intent $\gamma\in[0,1]$. Derived from retargeted demonstrations, $\gamma$ exposes aperture progression while retaining the full 6-D end-effector (EEF) and 6-DoF hand targets. Stage 1 predicts synchronized EEF--intent chunks. A conditional diffusion decoder then generates hand trajectories from these predictions, VLA context, and proprioception. Sequential optimization first establishes the EEF--intent interface, then learns hand control while allowing low-rate upstream adaptation.\par Evaluation separates reaching, stable grasp, and task completion to locate performance gains. With $\pi_{0.5}$, mean simulated task success rises from 50.3\% to 85.8\%, with a 51-point gain on Nut Assembly. Ablations, timing interventions, held-out objects, and hardware experiments test complementary aspects of the design.\par Our contributions are:\begin{enumerate}\item \textbf{A grasp-intent interface for dexterous VLA adaptation} that makes aperture progression explicit while preserving full supervision of embodiment-specific hand actions.\item \textbf{A coordinated generative policy} that combines synchronized EEF--intent prediction with conditional hand diffusion through sequential optimization and upstream co-adaptation.\item \textbf{A stage-resolved empirical evaluation} across tasks, backbones, held-out objects, and a physical platform, linking improved task completion to more reliable reach-to-grasp and grasp-to-task conversion.\end{enumerate}

\section{Related Work}

\subsection{Dexterous Manipulation}
Dexterous manipulation couples arm motion, hand morphology, and uncertain
contact. Studies of integrated humanoid platforms and hand representations
show that control abstractions cannot be separated from embodiment
~\cite{ref_schmitz2010humanoids,ref_grothe2012humanoids,ref_tomita2012humanoids}.
High-dimensional control under nonlinear contact exceeds parallel-gripper
closure~\cite{ref_bai2014,ref_chen2022}, while model-based and
reinforcement-learning approaches must additionally address dynamics, reward
design, and transfer~\cite{ref_kumar2014,ref_chen2022system,ref_zhao2020simtoreal}.
InDex targets a different boundary: the interface between pretrained VLA
priors and embodiment-specific joints. A low-dimensional cue represents
contact progress, but the decoder retains the full hand action rather than
reducing dexterity to binary open/close control.

\subsection{Teleoperation and Imitation Learning}
Teleoperation captures contact timing and finger coordination that are hard
to specify analytically. Humanoid systems have studied dexterous interfaces
and human-to-robot transfer~\cite{ref_to2012humanoids,ref_seo2023trill}, while
markerless vision reduces instrumentation~\cite{ref_anyteleop}.
Deterministic cloning can average multimodal demonstrations
~\cite{ref_pearce2023}; diffusion policies instead model action distributions
~\cite{ref_diffusionpolicy,ref_reuss2023}. InDex conditions such a decoder on
VLA features and retargeted intent, providing synchronized EEF, hand, and
transition supervision without separate event annotation.

\subsection{VLA Fine-tuning}
VLA models bind visual-language context to robot actions
~\cite{ref_rt2,ref_openvla,ref_pi05}, but efficient fine-tuning
~\cite{ref_lora} does not define a parallel-gripper-to-hand interface. Pan
\textit{et al.} switch from OpenVLA arm motion to a task-specific diffusion
grasp controller~\cite{ref_pan2024}. InDex instead retains a time-indexed
intent condition throughout synchronized EEF--hand chunks within one policy.
It is therefore a morphology-aware action interface, not a new foundation
model or a trigger between separately executed controllers.

\section{Method}

\begin{figure*}[t]
\centering
\includegraphics[width=0.96\textwidth]{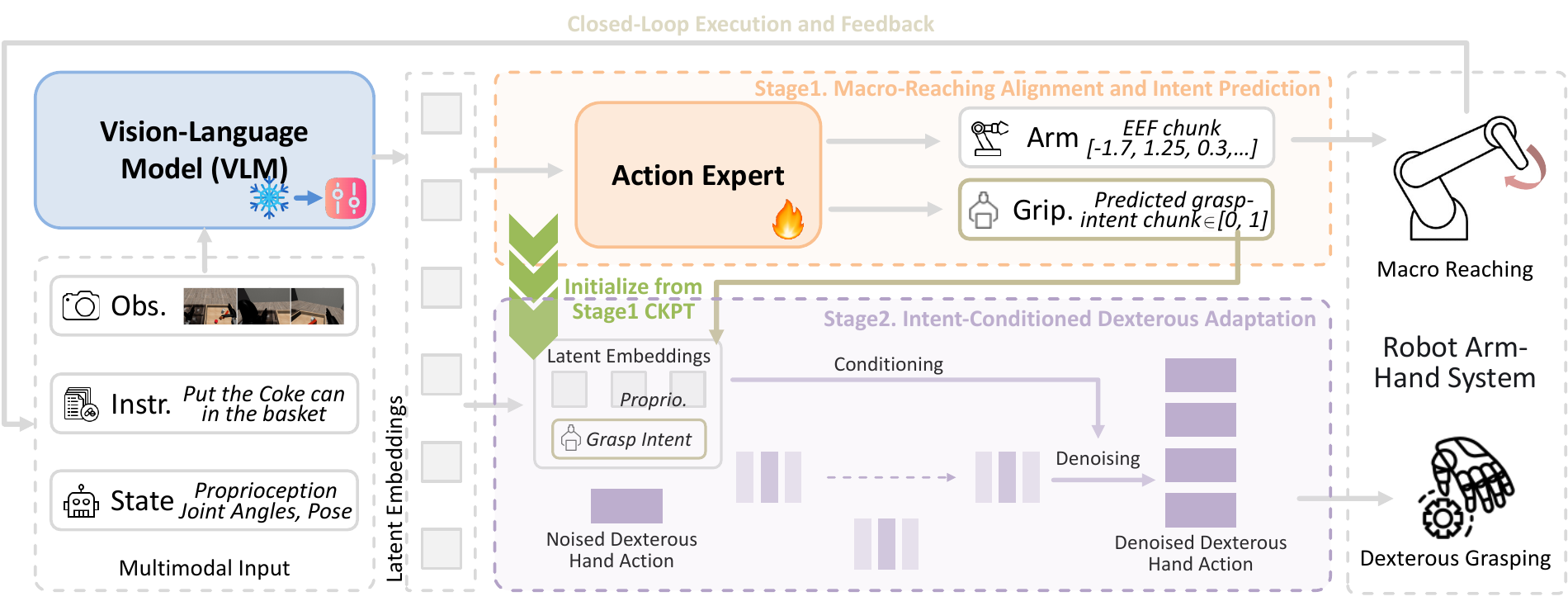}
    \caption{\textbf{InDex policy architecture.} Stage 1 predicts synchronized EEF--intent chunks. Conditioned on these predictions, VLA context, and proprioception, the Stage-2 diffusion decoder generates 6-DoF hand actions. EEF and hand actions form the executed 12-D command; intent is an auxiliary condition.}
	\label{fig:framework}
\end{figure*}

InDex links EEF motion to hand control through time-indexed grasp intent (Fig.~\ref{fig:framework}). We define the action interface and intent representation, then describe sequentially trained EEF--intent prediction and hand generation. Both modules operate together during deployment.

\subsection{Problem Formulation}
We learn an instruction-conditioned policy from demonstrations
$\mathcal{D}=\{(\mathbf{o}_i,\ell_i,\mathbf{a}_{eef,i},
\mathbf{a}_{hand,i},\gamma_i)\}_{i=1}^{N}$, where $\mathbf{o}_i$
contains global RGB, wrist RGB, and proprioception, and $\ell_i$ is the
language instruction. Each executed command comprises a 6-D Cartesian EEF
action $\mathbf{a}_{eef}\in\mathbb{R}^{6}$ and a 6-DoF hand action
$\mathbf{a}_{hand}\in\mathbb{R}^{6}$. The scalar $\gamma$, computed from the
retargeted hand state in Section~\ref{intent-realization}, conditions the decoder
but is not executed.
The EEF command is $\mathbf{a}_{eef}=[\Delta x,\Delta y,\Delta z,\Delta r_x,\Delta r_y,\Delta r_z]$, with translation and axis-angle rotation increments expressed in the robot base frame. At query time $t$, Stage 1 predicts $H$ synchronized EEF--intent pairs, $\{(\hat{\mathbf{a}}_{eef,t+j},\hat{\gamma}_{t+j})\}_{j=0}^{H-1}$. Stage 2 predicts the corresponding hand sequence, conditioned on these pairs, VLA context, and proprioceptive history. Training targets at each index come from the same demonstration timestamp. Concatenating EEF and hand predictions yields a chunk of $H$ 12-D commands; its prefix is executed before the next observation-driven query. Thus, intent tracks aperture progression within the chunk. Table~\ref{tab:training_protocol} specifies the horizons.

\subsection{Grasp-Intent Representation}
\label{intent-realization}
We compute $\gamma\in[0,1]$ from the robot hand configuration $\mathbf{q}$ obtained by retargeting (Section~\ref{teleop}). For the Fourier FDH-6 hand, the
joint state is
$\mathbf{q}=[\mathbf{q}_{th}^{\top},\mathbf{q}_{f}^{\top}]^{\top}
\in\mathbb{R}^{6}$. Let $p_{th}=\operatorname{FK}_{th}(\mathbf{q})\in\mathbb{R}^{3}$ and
$p_{f,i}=\operatorname{FK}_{f,i}(\mathbf{q})\in\mathbb{R}^{3}$ denote the thumb and
opposing-finger fingertips, and let
$\bar{p}_{f}=\frac{1}{4}\sum_{i=1}^{4}p_{f,i}$. We define the
\textit{virtual grasp aperture} as
\begin{equation}
d_{v}(\mathbf{q}) = \left\| p_{th} - \bar{p}_{f} \right\|_2,
\end{equation}
which we normalize using task- and embodiment-specific bounds. Here,
$e$ denotes the embodiment and $\tau$ the task.
\begin{equation}
\gamma =
\operatorname{clip}_{[0,1]}
\left(
\frac{d_{\max}^{(e,\tau)}-d_{v}(\mathbf{q})}
{d_{\max}^{(e,\tau)}-d_{\min}^{(e,\tau)}}
\right).
\end{equation}
Here, $\gamma=0$ and $1$ correspond to calibrated open and closed references. The Inspire-Robots hand uses its own forward kinematics and task-specific bounds. This many-to-one mapping standardizes aperture progression within each setting; it neither identifies contact directly nor uniquely specifies a hand pose. The decoder retains all six hand DoFs, using visual-language context and proprioception to resolve configurations compatible with the predicted EEF motion and intent.

\subsection{Stage 1: EEF--Intent Alignment}
\label{intent-prediction}
Our primary implementation uses $\pi_{0.5}$, whose vision-language model (VLM) conditions an Action Expert. Stage 1 retains the Cartesian EEF output and repurposes the parallel-jaw coordinate to predict $\gamma$, yielding $(\hat{\mathbf{a}}_{eef},\hat{\gamma})$ at each chunk index. The intent target is computed from demonstrated hand configurations using Section~\ref{intent-realization}. This interface preserves the distinction between arm motion and aperture progression while assigning finger-level control to Stage 2.

We apply LoRA~\cite{ref_lora} to the Action Expert and the staged VLM
schedule in Table~\ref{tab:training_protocol}. Unlike the Native baseline's
12-D flow-matching objective, Stage 1 separately supervises EEF translation
$\mathbf{p}$, rotation $\mathbf{r}$, and grasp intent:
\begin{equation}
\begin{aligned}
\mathcal{L}_{\mathrm{stage1}}
={}&\lambda_{pos}\mathcal{L}_{pos}
+\lambda_{rot}\mathcal{L}_{rot}
+\lambda_{intent}\mathcal{L}_{intent},\\
\mathcal{L}_{pos}
={}&\mathbb{E}\!\left[
\left\|\mathbf{p}^{*}-\hat{\mathbf{p}}\right\|_2^2
\right],\\
\mathcal{L}_{rot}
={}&\mathbb{E}\!\left[
\left\|\mathbf{r}^{*}-\hat{\mathbf{r}}\right\|_2^2
\right],\\
\mathcal{L}_{intent}
={}&\mathbb{E}\!\left[
\mathrm{BCE}_{\mathrm{logit}}(\hat{s}_{\gamma},\gamma^{*})
\right].
\end{aligned}
\end{equation}
Here, $\mathbf{a}_{eef}=[\mathbf{p}^{\top},\mathbf{r}^{\top}]^{\top}$,
$*$ denotes a target, and
$\hat{\gamma}=\operatorname{sigmoid}(\hat{s}_{\gamma})$.
$\mathrm{BCE}_{\mathrm{logit}}$ supports continuous
$\gamma^{*}\in[0,1]$. We set
$(\lambda_{pos},\lambda_{rot},\lambda_{intent})=(500,10,1)$.

\subsection{Stage 2: Conditional Hand Generation}
\label{dexterous-adaptation}
Stage 2 initializes the upstream modules from the final Stage-1 checkpoint and trains the hand decoder at $10^{-4}$. The VLM and Action Expert remain trainable at $10^{-6}$ (Table~\ref{tab:training_protocol}), allowing gradual co-adaptation of the learned EEF--intent interface.

The execution module is a conditional denoising diffusion model that
synthesizes multi-joint hand-action sequences $\mathbf{a}_{hand}$. For
compactness, let
$\mathbf{c}=[\mathbf{z}_{VLA},\mathbf{q}_{hist},
\hat{\mathbf{a}}_{eef},\hat{\gamma}]$ collect visual-language context,
proprioceptive history, and the synchronized Stage-1 prediction sequences.
Starting from $\mathbf{a}_{hand}^{K}\sim\mathcal{N}(0,I)$, a network
$\epsilon_\theta$ iteratively refines the sequence.
\begin{equation}
\begin{aligned}
\mathbf{a}_{hand}^{k-1}={}&
\alpha_k
\left[
\mathbf{a}_{hand}^{k}
-\eta_k\,\epsilon_\theta
(\mathbf{a}_{hand}^{k},\mathbf{c},k)
\right] \\
&+\sigma_k\boldsymbol{\xi}^{k},\\
\boldsymbol{\xi}^{k}\sim{}&\mathcal{N}(0,I).
\end{aligned}
\end{equation}
Here, $\alpha_k$, $\eta_k$, and $\sigma_k$ define the denoising schedule. We train the decoder with the noise-prediction objective:
\begin{equation}
\begin{aligned}
\tilde{\mathbf{a}}_{hand}^{k}={}&
\sqrt{\bar{\alpha}_k}\,\mathbf{a}_{hand}^{0}
+\sqrt{1-\bar{\alpha}_k}\,\boldsymbol{\epsilon}^{k}, \\
\mathcal{L}_{\mathrm{diff}}={}&
\mathbb{E}_{k,\boldsymbol{\epsilon}^{k}}
\!\left[
\left\|
\boldsymbol{\epsilon}^{k}-
\epsilon_\theta
(\tilde{\mathbf{a}}_{hand}^{k},\mathbf{c},k)
\right\|_2^2
\right].
\end{aligned}
\end{equation}
Here, $\mathbf{a}_{hand}^{0}$ is a demonstrated chunk, $\bar{\alpha}_k$ is the cumulative forward signal-retention factor, and $\boldsymbol{\epsilon}^{k}\sim\mathcal{N}(0,I)$. Gradients reach upstream modules through the predicted conditions. Both training and deployment use predicted EEF--intent sequences without oracle substitution. The decoder can model alternative hand trajectories, while the synchronized intent sequence provides a common reference for aperture progression.

\begin{figure*}[t]
\centering
\includegraphics[width=0.78\textwidth]{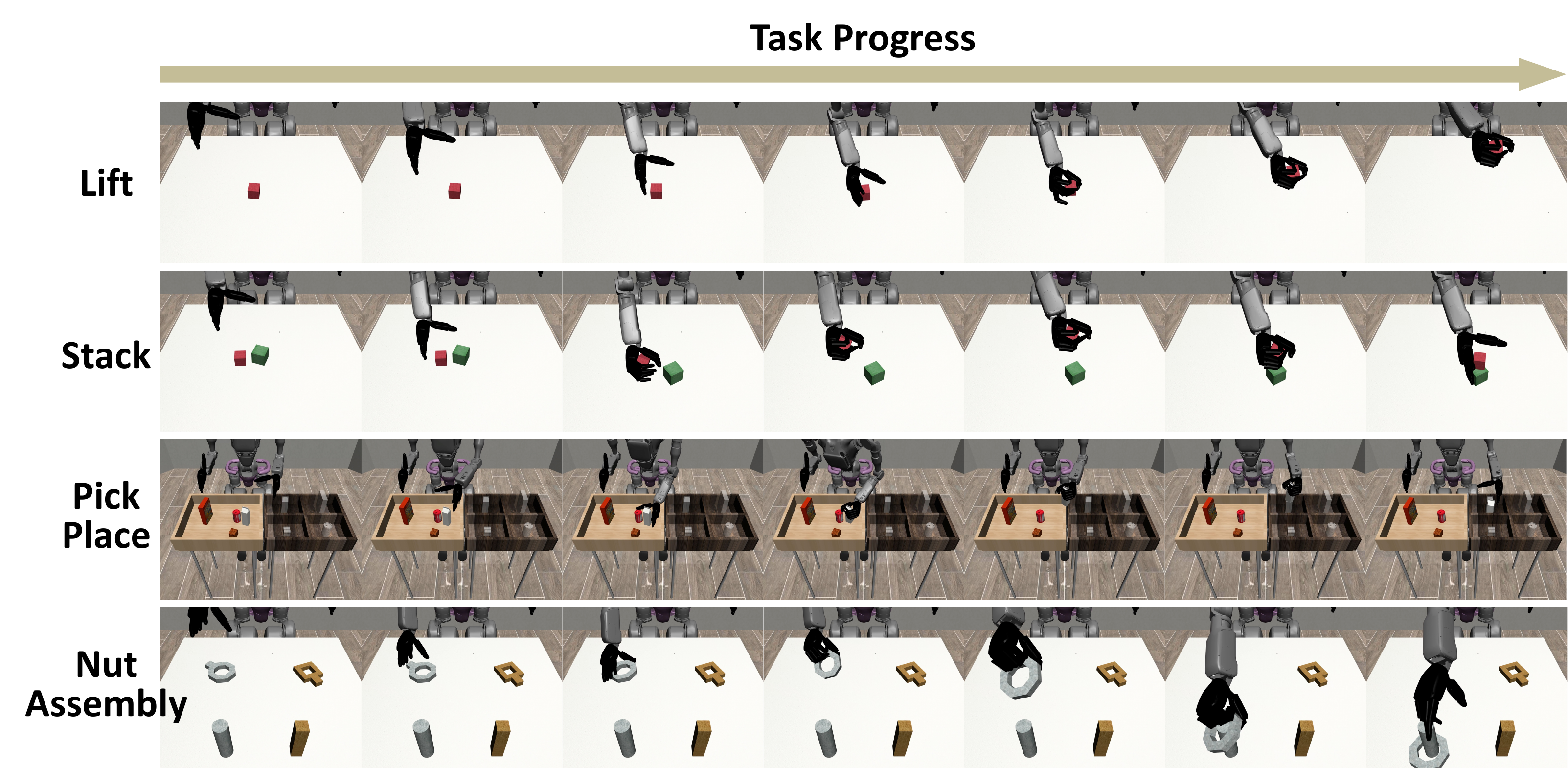}
    \caption{\textbf{Successful manipulation rollouts in simulation.} Time proceeds from left to right; rows show Lift, Stack, Pick \& Place, and Nut Assembly.}
	\label{fig:effectiveness_image}
\end{figure*}\section{Data and Platforms}
\label{data-platforms}
Operator measurements are first retargeted to each robot's 6-DoF hand space;
Section~\ref{intent-realization} then computes aperture and $\gamma$. Each
sample stores global and wrist RGB, proprioception, the 12-D action, and the
auxiliary cue. Embodiment-specific kinematics thus determine both the
supervised hand command and its intent condition.

\subsection{Unified Vision-Based Teleoperation}
\label{teleop}
An Intel RealSense D435i and AnyTeleop~\cite{ref_anyteleop} estimate hand keypoints and wrist pose, followed by retargeting under joint and smoothness constraints. Relative wrist motion sets EEF velocity, $\dot{\mathbf{p}}_{eef}=\mathbf{K}(\mathbf{w}_t-\mathbf{w}_{t-1})$; orientation maps as $\mathbf{R}_{eef}=\mathbf{R}_{cam}^{base}\mathbf{R}_{hand}\mathbf{R}_{align}$. Here, $\mathbf{w}_t\in\mathbb{R}^{3}$ is wrist position, $\mathbf{R}_{cam}^{base}$ the camera-to-base extrinsic, and $\mathbf{R}_{align}$ the anatomical-to-mechanical alignment. The operator camera is distinct from the policy cameras. Retargeted hand commands, computed intent, EEF motion, and observations share timestamps, preserving demonstrated arm--hand coordination.

\subsection{Simulation Platform and Corpus}
Simulation uses \textit{robosuite}~\cite{ref_robosuite} with a fixed-base Fourier GR-1 upper body and an FDH-6 hand. We collect 100 successful teleoperated demonstrations per task, recording global and wrist RGB, proprioception, 12-D EEF--hand commands, and $\gamma$ at 30 Hz in HDF5. Training and testing share the embodiment and observation interface; evaluation varies initial states and, in the held-out split, object instances.

\subsection{Simulation Tasks}
\label{simulation-tasks}
We adapt four \textit{robosuite} tasks to the Fourier arm--hand system:
\textit{Lift} raises a cube, \textit{Stack} places one cube on another,
\textit{Pick \& Place} transfers an object to its bin, and \textit{Nut
Assembly} seats a round nut on its peg. Success follows the corresponding
official benchmark criterion.

\subsection{Physical Platform and Corpus}
\label{physical-platform}
The physical platform combines a Rokae xMate ER3 Pro arm and a 6-DoF Inspire-Robots hand (Fig.~\ref{fig:real_success}, left). It uses the same D435i--AnyTeleop acquisition front end, with embodiment-specific workspace, EEF, retargeting, and aperture calibration. The policy receives global RGB, wrist RGB, and proprioception through the same EEF--intent--hand interface. The corpus covers \textit{Pick \& Place} and \textit{Block Fit}; the latter
aligns, inserts, and extracts an interlocking block within one trial. Each
task uses 50 successful demonstrations, and policies are trained on these
hardware data rather than transferred zero-shot. The real-world policy executes at 5 Hz, while a separate 1 kHz low-level controller tracks the selected set-points.

\begin{figure*}[t]
\centering
\includegraphics[width=0.94\textwidth]{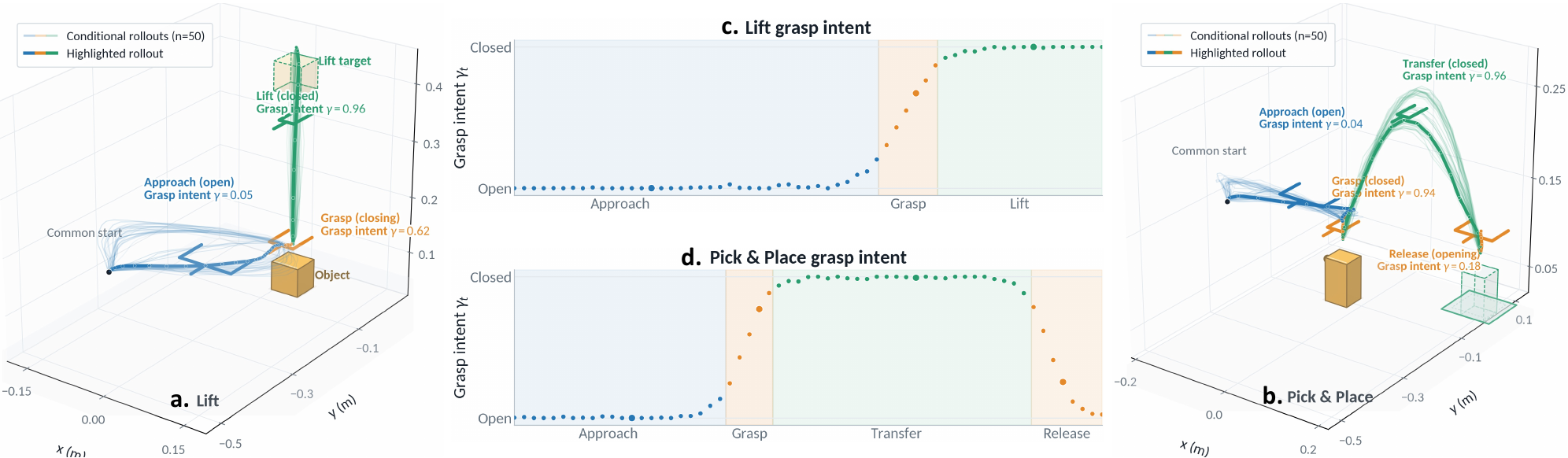}
    \caption{\textbf{Task-space rollout diversity and grasp-intent dynamics.} Translucent paths show 50 stochastic rollouts; solid paths highlight one execution. (a,b) Lift and Pick \& Place trajectories. (c,d) Intent traces for Lift (approach, grasp, elevation) and Pick \& Place (approach, grasp, transfer, release).}
    \label{fig:traj_intent_visual}
\end{figure*}\section{Experiments}
\label{experiments}
We compare simulation policies, ablate components, test backbone
compatibility, and separately evaluate a second physical embodiment.

\subsection{Evaluation Protocol}
\textbf{Baselines and training.} We compare BC-RNN~\cite{ref_mandlekar2021}, ACT~\cite{ref_zhao2023}, DP~\cite{ref_diffusionpolicy}, and native OpenVLA~\cite{ref_openvla}, UniVLA~\cite{ref_univla}, and $\pi_{0.5}$~\cite{ref_pi05}. Native VLAs retain their released decoder families, adapted to 12-D EEF--hand outputs, and train for 120k updates. InDex uses 60k updates per stage with batch size 128 (Table~\ref{tab:training_protocol}). Stage 1 freezes the VLM for 5k updates before unfreezing; Stage 2 reduces the upstream learning rate by $100\times$ relative to the hand decoder. \textit{Direct Projection} regresses actions from the final hidden state using a three-layer, 512-unit GELU MLP, MSE, and AdamW at $10^{-4}$. BC-RNN, ACT, DP, and Direct Projection train for 30k updates, at which their losses plateau; further training leads to overfitting.\par \textbf{Comparison controls.} Methods share demonstrations, observations, 12-D action targets, and evaluation initializations within each platform. InDex and native VLAs match total update counts, but architecture-dependent costs preclude a claim of matched FLOPs or GPU-hours. Training and inference use four RTX 4090 GPUs, each with 48 GB memory, and $640\times480$ RGB inputs.\par \textbf{Metrics and trial budgets.} Simulation reports cumulative success: $SR_{reach}$ requires EEF entry into a 3-cm object-centered sphere, $SR_{grasp}$ a secure hold for 30 simulator steps (1 s), and $SR_{task}$ the benchmark goal (Section~\ref{simulation-tasks}). Final checkpoints receive 100 simulated or 50 physical evaluation episodes per task, including failures in the denominators. These evaluation budgets are distinct from the 100/50 successful demonstrations used for training. Hardware execution and manual resets motivate the smaller physical budget; platform results remain separate. Rollout visualizations use 50 episodes, and timing interventions use 100 episodes per task and offset. Rates are computed from integer counts and rounded once to one decimal place.\par \textbf{Held-out object protocol.} For each simulated task, we substitute geometrically similar instances with different colors or models that are absent from training demonstrations. The trained policy and aperture bounds remain fixed, with no fine-tuning or recalibration. Each task receives 100 trials, yielding 400 per policy for each seen/unseen split. Equal task budgets make pooled success equivalent to the four-task mean. This tests object-instance variation within familiar task geometries. \begin{table}[t]
\caption{\textbf{InDex policy interface and optimization settings.}}
\label{tab:training_protocol}
\centering
\footnotesize
\setlength{\tabcolsep}{2.5pt}
\begin{tabularx}{\columnwidth}{@{}
>{\raggedright\arraybackslash}p{0.19\columnwidth}
>{\raggedright\arraybackslash}p{0.35\columnwidth}
>{\raggedright\arraybackslash}X@{}}
\toprule \textbf{Component} & \textbf{Parameter} & \textbf{Setting} \\ \midrule
Interface & Observation & Global/wrist RGB + proprio. \\
& Executed action & 6-D EEF + 6-DoF hand \\
& Auxiliary condition & Computed scalar $\gamma$ \\
\addlinespace[1.5pt]
Training & Batch size & 128 \\
& Reported checkpoint & Stage-2 final (60k) \\
\addlinespace[1.5pt]
Stage 1 & Training updates & 60k \\
& VLM schedule & Frozen: 0--5k; unfrozen: 5k--60k \\
& Trainable-module LR & $10^{-4}$ (VLM after 5k) \\
& LoRA $(r,\alpha)$ & $(16,32)$ \\
\addlinespace[1.5pt]
Stage 2 & Training updates & 60k \\
& Diffusion optimizer & AdamW \\
& Decoder learning rate & $10^{-4}$ \\
& VLM + Action Expert LR & $10^{-6}$ \\
\addlinespace[1.5pt]
Diffusion & Prediction horizon & 8 \\
& Observation history & 3 steps \\
& Executed actions/query & 6 \\
& Inference denoising steps & 100 \\
\addlinespace[1.5pt]
Compute & Training/inference GPUs & 4 $\times$ RTX 4090 (48 GB each) \\ \bottomrule
\end{tabularx}
\end{table}

\subsection{Simulation Evaluation}
\subsubsection{Main Multi-Task Performance}
\label{effectiveness}

\textbf{Overall performance.} Table~\ref{tab:main_results_final} compares cumulative success across methods and tasks. BC-RNN, ACT, and DP achieve 8.5\%, 34.5\%, and 42.8\% mean task success, respectively. Among native VLAs, $\pi_{0.5}$ performs best at 50.3\%, whereas InDex reaches 85.8\%. The improvement is 35.5 percentage points over native $\pi_{0.5}$ and 43.0 points over DP. Thus, the proposed adaptation improves on both task-trained imitation policies and pretrained VLA baselines.\par \textbf{Task difficulty.} InDex improves Lift, Stack, Pick \& Place, and Nut Assembly over native $\pi_{0.5}$ by 19, 40, 32, and 51 points, respectively. The larger gains on stacking and assembly indicate benefits beyond object approach. Nevertheless, InDex still drops from 95.0\% on Lift to 76.0\% on Nut Assembly, showing that precise post-grasp alignment remains challenging. Fig.~\ref{fig:effectiveness_image} illustrates the corresponding progression from approach and closure to transport, placement, or insertion. \begin{table*}[t]
\caption{\textbf{Multi-task manipulation performance in simulation.} SR denotes success rate. Entries are $SR_{reach}$ / $SR_{grasp}$ / $SR_{task}$. Each object-split aggregate pools 100 episodes per task (400 total); rates are rounded once to one decimal place.}
\label{tab:main_results_final}
\centering
\footnotesize
\setlength{\tabcolsep}{2.0pt}
\begin{tabular*}{\textwidth}{@{\extracolsep{\fill}}ccccccc@{}} \toprule \multirow{2}{*}{\textbf{Policy Variant}} & \multicolumn{6}{c}{\textbf{Cumulative Success Rate (\%)}} \\ \cmidrule(lr){2-7} & \textbf{Lift} & \textbf{Stack} & \makecell[c]{\textbf{Pick \&}\\\textbf{Place}} & \makecell[c]{\textbf{Nut}\\\textbf{Assembly}} & \makecell[c]{\textbf{Seen Objects}\\\textbf{Mean}} & \makecell[c]{\textbf{Unseen Objects}\\\textbf{Mean}} \\ \midrule BC-RNN & 48.0 / 31.0 / 24.0 & 21.0 / 5.0 / 1.0 & 33.0 / 15.0 / 9.0 & 12.0 / 3.0 / 0.0 & 28.5 / 13.5 / 8.5 & 12.5 / 5.5 / 2.0 \\ 
ACT & 78.0 / 65.0 / 61.0 & 53.0 / 33.0 / 27.0 & 65.0 / 45.0 / 41.0 & 35.0 / 15.0 / 9.0 & 57.8 / 39.5 / 34.5 & 29.8 / 19.0 / 18.0\\ 
DP & 83.0 / 73.0 / 68.0 & 59.0 / 43.0 / 37.0 & 71.0 / 55.0 / 51.0 & 45.0 / 21.0 / 15.0 & 64.5 / 48.0 / 42.8 & 50.5 / 32.0 / 22.5\\ 
\midrule OpenVLA (Native) & 85.0 / 69.0 / 63.0 & 41.0 / 23.0 / 17.0 & 59.0 / 41.0 / 34.0 & 46.0 / 19.0 / 13.0 & 57.8 / 38.0 / 31.8 & 49.8 / 35.0 / 25.8\\ 
UniVLA (Native) & 87.0 / 75.0 / 69.0 & 47.0 / 29.0 / 23.0 & 65.0 / 49.0 / 42.0 & 51.0 / 21.0 / 17.0 & 62.5 / 43.5 / 37.8 & 55.5 / 40.5 / 30.8\\ 
$\pi_{0.5}$ (Native) & 93.0 / 81.0 / 76.0 & 71.0 / 47.0 / 43.0 & 83.0 / 61.0 / 57.0 & 57.0 / 35.0 / 25.0 & 76.0 / 56.0 / 50.3 & 67.0 / 50.5 / 43.3\\ 
\midrule \textbf{$\pi_{0.5}$+InDex (Full)} & \textbf{98.0 / 97.0 / 95.0} & \textbf{91.0 / 86.0 / 83.0} & \textbf{95.0 / 91.0 / 89.0} & \textbf{87.0 / 79.0 / 76.0} & \textbf{92.8 / 88.3 / 85.8} & \textbf{82.5 / 73.3 / 69.0}\\ 
\bottomrule \end{tabular*}
\end{table*}

\begin{figure}[t]
\centering
\includegraphics[width=0.98\columnwidth]{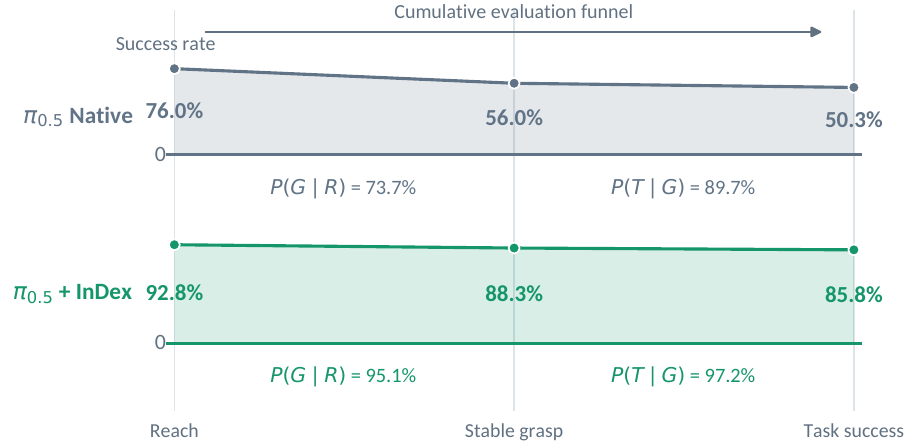}
    \caption{\textbf{Reach-to-grasp and grasp-to-task conversion.} Native $\pi_{0.5}$ and full InDex are compared using cumulative success rates and $P(G\mid R)$ and $P(T\mid G)$, pooled over 400 evaluation episodes per policy.}
    \label{fig:stage_conversion_funnel}
\end{figure}

\textbf{Stage-wise conversion.} Fig.~\ref{fig:stage_conversion_funnel} pools counts across the four tasks: $P(G\mid R)=N_G/N_R$ and $P(T\mid G)=N_T/N_G$. InDex improves reach from 76.0\% to 92.8\%, reach-to-grasp conversion from 73.7\% to 95.1\%, and grasp-to-task conversion from 89.7\% to 97.2\%. Of 371 reached episodes, 353 establish a stable grasp and 343 complete the task; native $\pi_{0.5}$ obtains 304, 224, and 201. The gains therefore reflect both better approach and fewer failures after reaching.\par \textbf{Held-out objects.} InDex achieves 69.0\% task success versus 43.3\% for native $\pi_{0.5}$, a 25.8-point gain computed before rounding. Its reach/grasp/task rates of 82.5/73.3/69.0\% remain the highest among compared methods. However, its seen-to-unseen task-success drop is 16.8 points, compared with 7.0 for the native policy. Thus, InDex retains a substantial absolute advantage but does not eliminate sensitivity to object-instance changes.\par \textbf{Trajectory and intent dynamics.} Fig.~\ref{fig:traj_intent_visual} shows 50 stochastic rollouts per visualized task from a common nominal initialization, with one highlighted. Spatial paths vary while intent follows the demonstrated phase structure: it rises around grasping and remains high during Lift, then decreases for release in Pick \& Place. This is qualitatively consistent with a shared temporal cue supporting varied trajectories; it does not by itself measure hand-trajectory distribution coverage. \subsubsection{Ablation Study}\label{ablation}\begin{figure}[t]
\centering
\includegraphics[width=0.98\columnwidth]{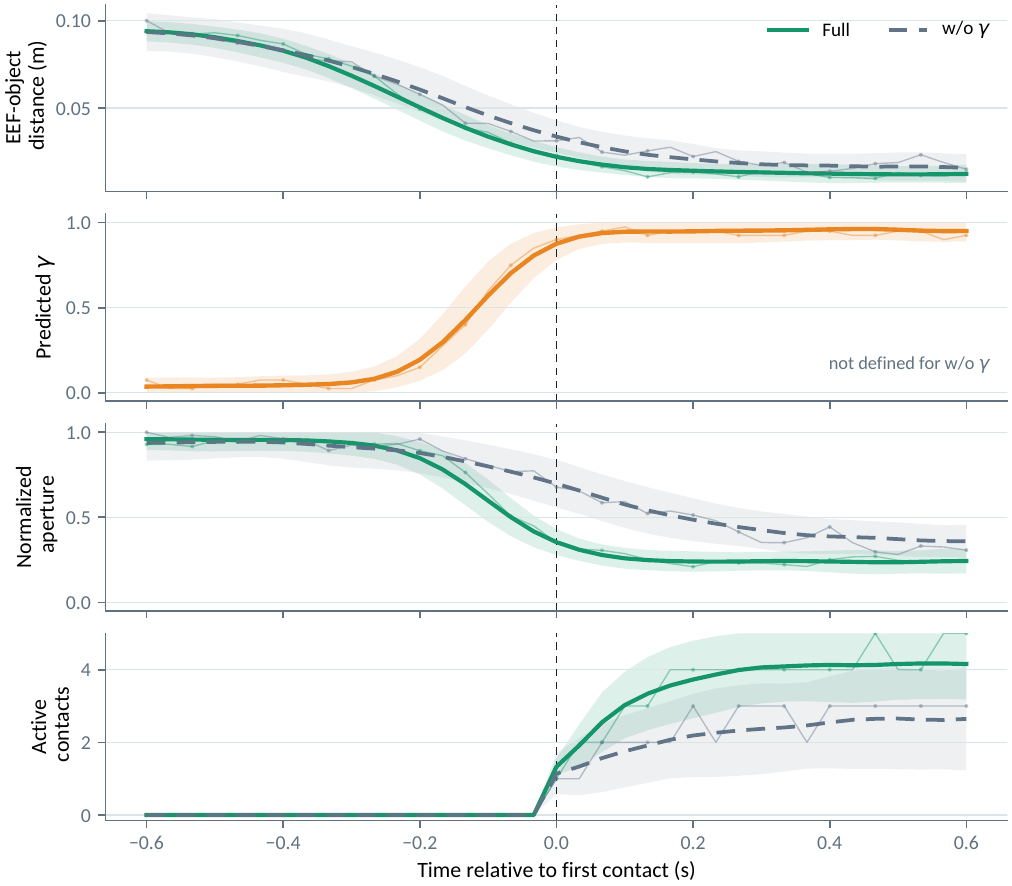}
    \caption{\textbf{Contact-aligned arm--hand coordination on Lift.} Across 50 rollouts, time zero denotes first contact. Curves show smoothed means, bands show $\pm2$ smoothed standard deviations, and point-connected traces show one rollout.}
    \label{fig:contact_aligned_coordination}
\end{figure}\begin{figure}[t]
\centering
\includegraphics[width=0.98\columnwidth]{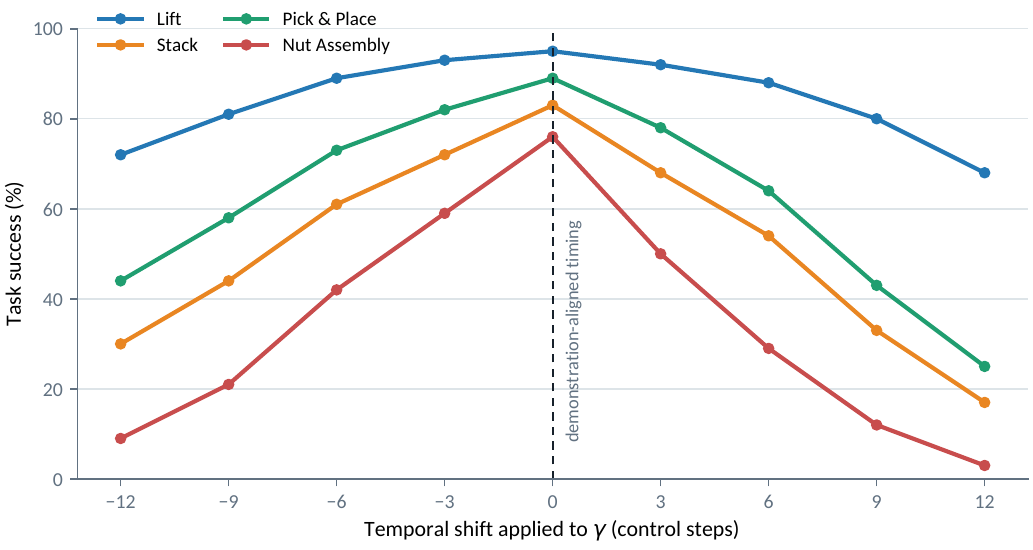}
    \caption{\textbf{Sensitivity to grasp-intent timing.} Negative offsets advance and positive offsets delay $\gamma$ while EEF predictions remain unchanged. Each point reports task success over 100 evaluation episodes.}
    \label{fig:gamma_timing_sensitivity}
\end{figure}\begin{figure*}[!t]
    \centering
    \includegraphics[width=0.78\textwidth]{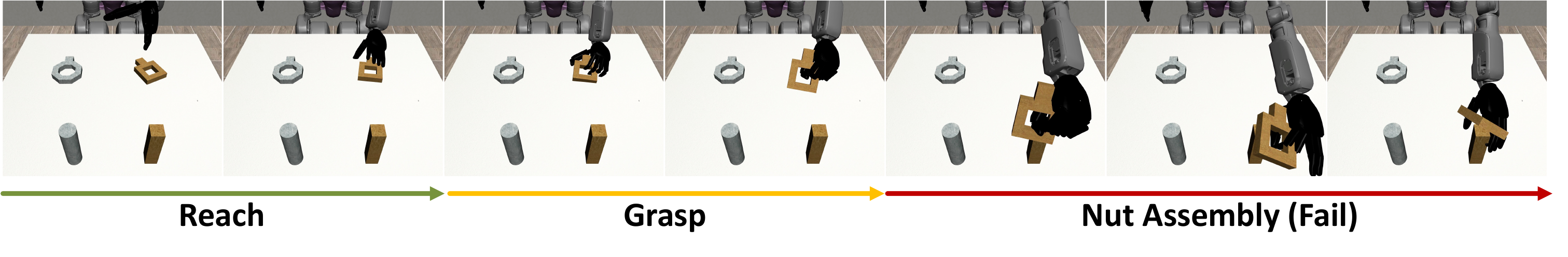}
    \\[0.05cm]
    \includegraphics[width=0.78\textwidth]{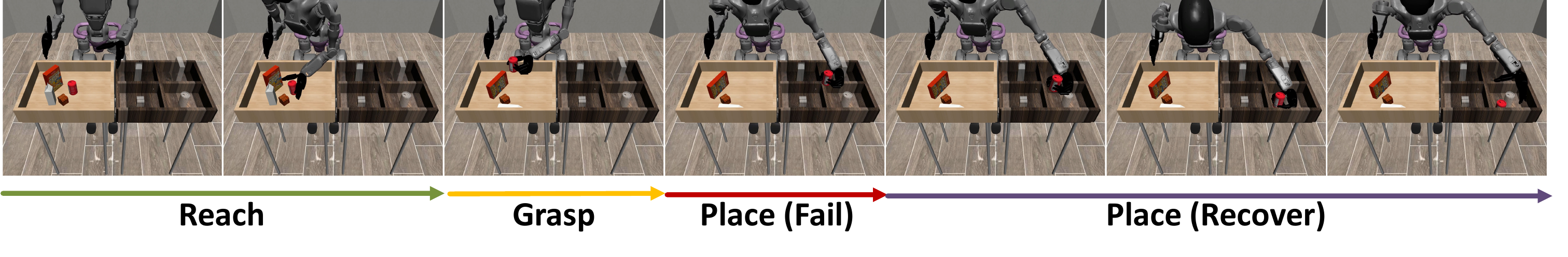}
    \caption{\textbf{Failure modes and closed-loop recovery.} Time proceeds from left to right. Top: a Nut Assembly failure after reaching. Bottom: Pick \& Place recovery from an initially tilted object.}
    \label{fig:failure}
\end{figure*}\begin{figure*}[!t]
    \centering
    \includegraphics[width=0.90\textwidth]{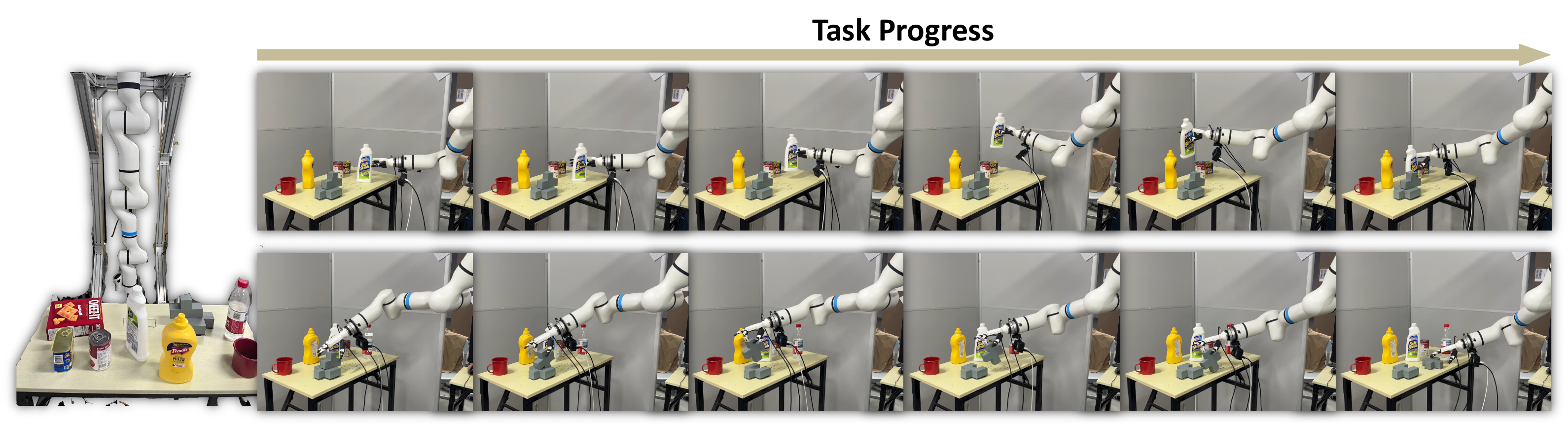}
    \caption{\textbf{Real-world platform and manipulation rollouts.} Left: the Rokae xMate ER3 Pro--Inspire-Robots platform. Right: successful Pick \& Place and Block Fit rollouts, with time proceeding from left to right.}
    \label{fig:real_success}
\end{figure*}

Table~\ref{tab:ablation_study} tests four variants: Vision-only removes the broader VLA-derived conditioning pathway; w/o $\gamma$ removes only intent prediction and conditioning; Coupled-12k omits Stage-1 initialization; and MLP Hand replaces diffusion with a 256--512--1024--512--256 MLP. Conditional rates locate failures within the execution sequence.

\begin{table}[htbp]
\centering
\caption{\textbf{Policy component ablations in simulation.}}
\label{tab:ablation_study}

\scriptsize
\setlength{\tabcolsep}{2.0pt}

\begin{tabularx}{\columnwidth}{@{}
>{\centering\arraybackslash}p{0.25\columnwidth}
*{5}{>{\centering\arraybackslash}X}
*{2}{>{\centering\arraybackslash}p{0.13\columnwidth}}@{}}
\toprule \multirow{2}{*}{\textbf{Policy Variant}} & \multicolumn{4}{c}{\textbf{Task SR (\%)}} & \multirow{2}{*}{\makecell[c]{\textbf{Mean}\\\textbf{(\%)}}} & \multicolumn{2}{c}{\textbf{Cond. SR (\%)}} \\ \cmidrule(lr){2-5}\cmidrule(lr){7-8} & \textbf{Lift} & \textbf{Stack} & \textbf{P\&P} & \textbf{Nut} & & {$P(G\!\mid\!R)$} & {$P(T\!\mid\!G)$} \\ \midrule

\mbox{$\pi_{0.5}$ Native} & 76.0 & 43.0 & 57.0 & 25.0 & 50.3 & 73.7 & 89.7 \\ \midrule
\mbox{Direct Proj.}       & 13.0 & 0.0  & 3.0  & 0.0  & 4.0 &
13.2 & 66.7 \\
\mbox{InDex: Vision-only} & 37.0 & 12.0 & 19.0 & 0.0  & 17.0 &
33.4 & 68.0 \\ \midrule

\mbox{InDex: Coupled-12k} & 45.0 & 14.0 & 22.0 & 5.0  & 21.5 &
42.4 & 67.2 \\
\mbox{InDex: MLP Hand}   & 68.0 & 42.0 & 55.0 & 25.0 & 47.5 &
82.0 & 65.1 \\
\mbox{InDex: w/o $\gamma$} & 82.0 & 58.0 & 66.0 &
34.0 & 60.0 & 75.6 & 88.9 \\ \midrule

\mbox{\textbf{InDex: Full}} & \textbf{95.0} & \textbf{83.0} & \textbf{89.0} & \textbf{76.0} & \textbf{85.8} & \textbf{95.1} & \textbf{97.2} \\ \bottomrule
\end{tabularx}
\par\vspace{3pt}\noindent
\parbox{\columnwidth}{\scriptsize $R$, $G$, and $T$ denote reach, stable
grasp, and task completion; $P(G\mid R)$ and $P(T\mid G)$ are pooled
conditional success rates.}
\end{table}

\textbf{Conditioning and training controls.} Direct Projection and Vision-only achieve 4.0\% and 17.0\% mean success, respectively, showing that these simplified interfaces are insufficient in this setting. Vision-only removes multiple conditioning signals and cannot identify the effect of $\gamma$ alone. Coupled-12k reaches 21.5\%, but simultaneously changes initialization, learning rates, and training duration; it compares training configurations without isolating sequential optimization.\par \textbf{Intent and hand generation.} The targeted w/o $\gamma$ control reduces mean success from 85.8\% to 60.0\%, with task-wise losses of 13, 25, 23, and 42 points. Its reach-to-grasp conversion falls from 95.1\% to 75.6\%, supporting the value of the intent prediction-and-conditioning pathway. Replacing diffusion with MLP Hand lowers mean success to 47.5\%; although 82.0\% of reaches yield a grasp, only 65.1\% of grasps complete the task, versus 97.2\% for Full. The two controls therefore expose complementary weaknesses in contact establishment and subsequent manipulation. They support the chosen cue and decoder on these tasks, without uniquely attributing all decoder gains to multimodality. \par \textbf{Contact-aligned behavior.} Fig.~\ref{fig:contact_aligned_coordination} aligns 50 Lift rollouts per variant at 30 Hz to first contact ($t=0$). It shows five-sample Gaussian-smoothed means, $\pm2$ smoothed standard-deviation bands, and one point-connected rollout. Full raises $\gamma$ during late approach and maintains it during elevation; closure precedes contact accumulation. Without $\gamma$, approach is similar but closure is delayed and fewer contacts form, consistent with the reach-to-grasp gap. Bands indicate rollout variability, not confidence intervals.\par \textbf{Timing intervention.} Fig.~\ref{fig:gamma_timing_sensitivity} shifts $\gamma$ by $\{-12,-9,\ldots,12\}$ control steps while retaining EEF predictions and the same 100 initial conditions per task and offset. At 30 Hz, this spans $\pm0.4$ s in 0.1-s increments. Success peaks near the unmodified policy (zero offset): Lift tolerates larger shifts, Stack and Pick \& Place degrade faster, and Nut Assembly is especially delay-sensitive. This intervention supports a contribution from temporal alignment beyond the presence of intent alone.\par \subsubsection{Failure and Recovery Analysis}

Fig.~\ref{fig:failure} shows Nut Assembly stalling after reaching the insertion region, while Pick \& Place recovers from an initially tilted object through subsequent observation-conditioned chunks. These examples illustrate residual alignment errors and limited closed-loop correction, without establishing general disturbance recovery. \subsubsection{Cross-Backbone Evaluation}
\label{transferability}
Table~\ref{tab:transferability_analysis} compares variants sharing the pretrained checkpoint and evaluation protocol within each backbone. InDex improves native mean success by 14.5, 14.8, and 35.5 points for OpenVLA, UniVLA, and $\pi_{0.5}$, respectively, with gains on every task. Stack gains are 21/21/40 points and Nut Assembly gains 12/14/51 points, whereas Direct Projection averages only 2.5--4.0\%. These results support cross-backbone compatibility, while unequal gains indicate dependence on upstream representations and optimization. \begin{table}[htbp]
\centering
\caption{\textbf{Cross-backbone policy evaluation in simulation.}}
\label{tab:transferability_analysis}

\footnotesize
\setlength{\tabcolsep}{1.5pt}

\begin{tabularx}{\columnwidth}{@{}>{\centering\arraybackslash}p{0.44\columnwidth}*{5}{>{\centering\arraybackslash}X}@{}}
\toprule \multirow{2}{*}{\textbf{Policy Variant}} & \multicolumn{4}{c}{\textbf{Task SR (\%)}} & \multirow{2}{*}{\makecell[c]{\textbf{Mean}\\\textbf{(\%)}}} \\ \cmidrule(lr){2-5} & \textbf{Lift} & \textbf{Stack} & \textbf{P\&P} & \textbf{Nut} & \\ \midrule

\mbox{OpenVLA (Direct Proj.)} & 8.0 & 0.0 & 2.0 & 0.0 & 2.5 \\
\mbox{OpenVLA (Native)} & 63.0 & 17.0 & 34.0 & 13.0 & 31.8 \\
\mbox{\textbf{OpenVLA + InDex}} & \textbf{73.0} & \textbf{38.0} & \textbf{49.0} & \textbf{25.0} & \textbf{46.3} \\ \midrule

\mbox{UniVLA (Direct Proj.)} & 11.0 & 0.0 & 4.0 & 0.0 & 3.8 \\
\mbox{UniVLA (Native)} & 69.0 & 23.0 & 42.0 & 17.0 & 37.8 \\
\mbox{\textbf{UniVLA + InDex}} & \textbf{79.0} & \textbf{44.0} & \textbf{56.0} & \textbf{31.0} & \textbf{52.5} \\ \midrule

\mbox{$\pi_{0.5}$ (Direct Proj.)} & 13.0 & 0.0 & 3.0 & 0.0 & 4.0 \\
\mbox{$\pi_{0.5}$ (Native)} & 76.0 & 43.0 & 57.0 & 25.0 & 50.3 \\
\mbox{\textbf{$\pi_{0.5}$ + InDex (Full)}} & \textbf{95.0} & \textbf{83.0} & \textbf{89.0} & \textbf{76.0} & \textbf{85.8} \\ \bottomrule
\end{tabularx}
\end{table}

\subsection{Physical-Robot Evaluation}
\label{physical-evaluation}
We evaluate DP and the Native and InDex variants of UniVLA and $\pi_{0.5}$
on the Rokae--Inspire platform. Policies share 50 demonstrations,
observations, 12-D actions, test initializations, and 50 trials per task.
Table~\ref{tab:hardware_results} reports final-checkpoint success, and
Fig.~\ref{fig:real_success} shows the platform and representative executions.

\begin{table}[htbp]
\centering
\caption{\textbf{Real-world task performance (50 episodes per task).}}
\label{tab:hardware_results}
\footnotesize
\setlength{\tabcolsep}{2pt}
\begin{tabularx}{\columnwidth}{@{}
>{\centering\arraybackslash}p{0.40\columnwidth}
*{3}{>{\centering\arraybackslash}X}@{}}
\toprule \multirow{2}{*}{\textbf{Policy Variant}} & \multicolumn{2}{c}{\makecell[c]{\textbf{Successful / Total}\\\textbf{Evaluation Episodes}}} & \multirow{2}{*}{\makecell[c]{\textbf{Mean}\\\textbf{(\%)}}} \\ \cmidrule(lr){2-3} & \makecell[c]{\textbf{Pick \&}\\\textbf{Place}} & \textbf{Block Fit} & \\ \midrule
DP & $25 / 50$ & $22 / 50$ & 47.0 \\
UniVLA (Native) & $29 / 50$ & $25 / 50$ & 54.0 \\
UniVLA+InDex & $38 / 50$ & $32 / 50$ & 70.0 \\
$\pi_{0.5}$ (Native) & $40 / 50$ & $31 / 50$ & 71.0 \\
\textbf{$\pi_{0.5}$+InDex} & $46 / 50$ & $37 / 50$ &
\textbf{83.0} \\ \bottomrule
\end{tabularx}
\end{table}

Table~\ref{tab:hardware_results} shows mean success increasing from 54.0\% to 70.0\% for UniVLA and from 71.0\% to 83.0\% for $\pi_{0.5}$, versus DP's 47.0\%. InDex adds six successes per task to $\pi_{0.5}$, reaching 92.0\% on Pick \& Place and 74.0\% on Block Fit. The remaining 18-point gap is consistent with the difficulty of precise fitting. Fig.~\ref{fig:real_success} illustrates transport/release and insertion/extraction. These results support deployment with embodiment-specific training, not zero-shot sim-to-real transfer. \subsection{Discussion and Limitations} Update-matched VLA comparisons do not establish compute-matched efficiency. Isolating sequential training requires joint training with matched initialization, learning rates, and budget. Aperture intent may be insufficient for independent contacts, finger gaiting, or in-hand rotation; without tactile feedback, recovery is restricted to visually or proprioceptively observable errors. \section{Conclusion} InDex couples synchronized grasp-intent prediction with conditional hand generation to adapt VLA priors to dexterous control. Task-level, stage-wise, and intervention results support improved arm--hand coordination across three backbones and separately trained hardware policies. The intent pathway is supported by a targeted ablation; sequential training still requires matched controls. Generalization is demonstrated only for geometrically similar held-out objects within known tasks. Aperture calibration remains task- and embodiment-specific; unseen-task and unseen-hand transfer are unverified. Other 6-DoF hands preserve the output dimension but still require retargeting, calibration, and training; different dimensions require decoder adaptation. Future work will investigate object-conditioned latent aperture normalization, richer contact representations, tactile feedback, and broader transfer. \balance\section*{ACKNOWLEDGMENT}

This work was supported in part by the National Key Research and Development Program of China under Grant 2023YFB4704305; in part by the National Natural Science Foundation of China under Grant U22B2080; in part by the Open Foundation of the National Key Laboratory of Multi-perch Vehicle Propulsion Systems; and in part by the Beijing Nova Program.

\bibliographystyle{IEEEtran}       % 指定使用 IEEEtran 格式
\bibliography{IEEEabrv, references} % 引入 bib 文件（注意不写 .bib 后缀）

\end{document}